\documentclass{article}
\usepackage{iclr2026_conference,times}

\usepackage{amsmath,amsfonts,bm}

\def\eqref#1{equation~\ref{#1}}

\def\1{\bm{1}}

\DeclareMathAlphabet{\mathsfit}{\encodingdefault}{\sfdefault}{m}{sl}
\SetMathAlphabet{\mathsfit}{bold}{\encodingdefault}{\sfdefault}{bx}{n}

\usepackage{hyperref}
\usepackage{url}
\usepackage{cleveref}
\usepackage{orcidlink}

\title{RocqSmith: Can Automatic Optimization \\ Forge Better Proof Agents?}

\author{
    Andrei Kozyrev$^{1}$, Nikita Khramov$^{1}$, Denis Lochmelis$^{1}$, Valerio Morelli$^{1}$ \\
    ~\textbf{Gleb Solovev$^{1}$, Anton Podkopaev$^{2,3}$} \\
    ~$^1$JetBrains Research Germany, $^2$JetBrains Research Netherlands, \\
    ~$^3$Constructor University Bremen
}

\usepackage[colorinlistoftodos,textsize=tiny]{todonotes}

\crefname{part}{\S}{\S\S}
\crefname{chapter}{\S}{\S\S}
\crefname{section}{\S}{\S\S}
\crefname{subsection}{\S}{\S\S}
\crefname{figure}{fig.}{Fig.}

\iclrpreprintcopy
\begin{document}

\maketitle
\begin{abstract}
This work studies the applicability of automatic AI agent optimization methods to real-world agents in formal verification settings, focusing on automated theorem proving in Rocq as a representative and challenging domain. We evaluate how different automatic agent optimizers perform when applied to the task of optimizing a Rocq proof-generation agent, and assess whether parts of the fine-grained tuning of agentic systems, such as prompt design, contextual knowledge, and control strategies, can be automated. Our results show that while several optimizers yield measurable improvements, simple few-shot bootstrapping is the most consistently effective; however, none of the studied methods matches the performance of a carefully engineered state-of-the-art proof agent.
\end{abstract}

\section{Introduction}

Recent advances in Generative Artificial Intelligence (AI) have significantly accelerated software development workflows. At the same time, empirical studies indicate that code generated or assisted by large language models may still exhibit security weaknesses and correctness issues in real-world software projects~\citep{fu2025weaknessesofcoqpilot}. Formal methods provide a way to mitigate these risks by offering strong correctness guarantees and explicit acceptance criteria for generated software, assuming a well-formed specification. Interactive Theorem Provers (ITPs) support the development of formal specifications and machine-checked proofs. Several mature ITPs exist today, including Rocq (formerly Coq)~\citep{coq}, Lean~\citep{lean}, Agda~\citep{agda}, and Isabelle~\citep{isabelle}. Among them, Rocq is a particularly mature system with over 30 years of continuous development and a strong track record in high-impact verification efforts, including the formal verification of the CompCert C compiler~\citep{compcert}, the only widely used C compiler for which an extensive study reported no miscompilation bugs~\citep{yang2011finding}.

Formal software verification remains a demanding process that requires substantial human expertise, motivating extensive work on automating theorem proving in Rocq. One promising direction is the development of autonomous agentic systems for interactive theorem proving~\citep{xu2026agentic}. Agentic systems are a natural fit for this setting, as their interaction loop closely resembles that of a human Rocq programmer. An agent observes the current environment state, performs intermediate reasoning, and selects an action to apply. In Rocq, proof construction follows an analogous process: the user (or agent) iteratively inspects the current \emph{proof state}, consisting of the active goal and its context, and applies a \emph{tactic} that transforms the proof state into simpler subgoals until the proof is complete. The classical \emph{ReAct} (Reason-and-Act) paradigm~\citep{react} aligns well with this workflow, alternating between reasoning about the proof state and selecting tactics to apply in a step-by-step manner.

Several attempts have been made to build autonomous agentic systems for interactive theorem proving~\citep{yang2023leandojo, nlir, kozyrev2025rocqstar}. Among them, RocqStar~\citep{kozyrev2025rocqstar} proposes an agentic system for generating Rocq proofs and reports strong empirical results on the IMM-300 benchmark. The RocqStar agent consists of a multi-component pipeline that combines planning, execution, and auxiliary control mechanisms, and relies on a substantial amount of carefully engineered prompts and agent interactions.
Although such systems demonstrate the potential of agentic approaches to theorem proving, their design and implementation involve considerable engineering effort. Constructing and tuning a production-grade proof agent typically requires manual prompt design, task decomposition, tool orchestration, and the curation of structured contextual knowledge. This motivates the study of whether automatic agent optimization methods can reduce the amount of manual engineering required to build and adapt such agents.

Recent work on automatic agent optimization aims to reduce manual agent engineering by placing agents into feedback-driven optimization loops. Depending on the approach, optimization may target prompts and few-shot demonstrations~\citep{mipro}, external contextual knowledge~\citep{reasoningbank, ace}, or the agent’s control flow structure~\citep{adas}. Most of the existing evaluations of automatic agent optimization methods focus on benchmark-driven tasks such as SWE-Bench or AppWorld. However, it remains unclear how these methods generalize to agents in formal domains where correctness is strictly verified rather than empirically tested. Interactive theorem proving serves as a representative example of such a task; it provides a strong verifier that allows evaluation of the agent's actions and returns a clear feedback signal from the environment, while at the same time posing a challenging problem that cannot be solved in an ad hoc manner and requires structured reasoning. In this work, we take a simplified but representative core of a production-grade Rocq proof agent and use it as a test case to evaluate automatic agent optimization methods. We systematically study how different optimizers perform when applied to this setting and assess their ability to improve the agent toward the performance level of a carefully human-designed baseline.

\section{Preliminaries}\label{sec:preliminaries}

RocqStar~\citep{kozyrev2025rocqstar} is an agentic system for Rocq proof generation built as a multi-component pipeline that combines planning, execution, reflection, critique, and retrieval-based replanning, and relies on carefully constructed prompts, auxiliary control logic, and a rich tool interface for interacting with the prover. Although this design achieves strong empirical performance, it requires substantial manual engineering effort to construct, tune, and maintain. In this work, we omit these hand-crafted components and consider a simplified agent core as our baseline. In particular, we study a single-loop agent topology, where a minimal control structure repeatedly observes the current proof state, performs lightweight reasoning, selects a tool call, and executes it until termination. This simplified agent captures the essential interaction pattern between an LLM and the Rocq environment while avoiding additional planning or control modules. Our goal is to evaluate whether automatic agent optimization methods can bridge the performance gap between such a minimal agent and a carefully engineered, human-designed agentic pipeline.

We begin our study by evaluating the automatic agent optimization methods available in the DSPy framework~\citep{dspy}. We implement a simplified RocqStar agent in DSPy and apply a range of built-in optimizers, including \emph{BootstrapFewShot}, \emph{MIPROv2}~\citep{mipro}, \emph{SIMBA}, and \emph{GEPA}~\citep{gepa}. \emph{BootstrapFewShot} constructs few-shot prompts by collecting successful execution traces on a training set and including the first $k$ (4 in evaluation due to context limitations) of such demonstrations directly into the prompt, providing a simple baseline for in-context learning. \emph{MIPROv2} jointly optimizes instructions and few-shot demonstrations by proposing multiple instruction candidates and performing a discrete search over their combinations using Bayesian optimization~\citep{bayesianoptimization} guided by task performance. \emph{SIMBA} applies a random search over prompt variants, keeping changes that lead to improved task performance. \emph{GEPA} applies an evolutionary optimization strategy to the textual components of the agent, such as prompts or instructions, using selection mechanisms and optional reflective feedback to progressively improve performance.

In addition to prompt-centric optimizers, we consider approaches that optimize an agent’s contextual knowledge based on past behavior. Both \emph{ACE}~\citep{ace} and \emph{ReasoningBank}~\citep{reasoningbank} operate by running the agent on a training dataset, analyzing successful and failed executions using a language model, and storing the resulting reflections in a structured knowledge base that can be reused at test time. This knowledge captures high-level guidance about effective strategies and common failure modes. The two approaches differ primarily in how this contextual knowledge is represented and retrieved. \emph{ReasoningBank} maintains a memory bank of past experiences and uses an encoder model to embed the current query at test time and retrieve the most similar memory items, which are then injected into the agent’s context. \emph{ACE} instead maintains a curated, agent-specific context that is incrementally refined across training iterations, and provides the resulting context to the agent in full at inference time in order to better utilize the available context window, rather than relying on similarity-based retrieval. Finally, we also evaluate \emph{ADAS}~\citep{adas}, which targets the agent’s control flow rather than its prompts or context. ADAS represents the agent as executable code and optimizes its topology by prompting a language model with the available tools and task description to generate code defining the agent’s overall decision logic.

\section{Experiments}

\begin{table*}[t]
\centering
\small
\begin{tabular}{lllp{3.5cm}c|ccc|c}
\hline
\textbf{Impl} & \textbf{Opt. Time} & \textbf{Agent} & \textbf{Description} & \textbf{Model} &
\multicolumn{3}{c|}{\textbf{Proof Length}} & \textbf{Total} \\
 &  &  &  &  &
$\leq$4 & 5--8 & 9--20 &  \\
\hline
Koog & --- & Simple & --- & GPT-4.1 & 10\% & 10\% & 0\% & 6\% \\
Koog & 45 min & Simple & w/ BootstrapFewShot & GPT-4.1 & 24\% & 10\% & 6\% & 13\% \\
Koog & 3h & Simple & w/ ACE & GPT-4.1 & 22\% & 6\% & 2\% & 10\% \\
Koog & 2h & Simple & w/ RB & GPT-4.1 & 32\% & 14\% & 4\% & 17\% \\
Koog & --- & ReAct & --- & GPT-4.1 & 34\% & 16\% & 10\% & 20\% \\
Koog & 45 min & ReAct & w/ BootstrapFewShot & GPT-4.1 & 54\% & 28\% & 18\% & \textbf{33\%} \\
Koog & 6h & ReAct & w/ ACE & GPT-4.1 & 28\% & 16\% & 6\% & 17\% \\
Koog & 6h & ReAct & w/ RB & GPT-4.1 & 40\% & 22\% & 12\% & 25\% \\
Koog & 6h & ReAct & w/ RB (no retrieval) & GPT-4.1 & 32\% & 20\% & 12\% & 21\% \\
\hline
ADAS & 70h & ADAS & --- & GPT-4.1 & 30\% & 14\% & 8\% & 17\% \\
\hline
DSPy & --- & ReAct & --- & GPT-4.1 & 38\% & 16\% & 4\% & 19\% \\
DSPy & 45 min & ReAct & w/ BootstrapFewShot & GPT-4.1 & 66\% & 32\% & 22\% & 40\% \\
DSPy & 12h & ReAct & w/ MIPROv2 & GPT-4.1 & 66\% & 42\% & 22\% & \textbf{43\%} \\
DSPy & 14h & ReAct & w/ SIMBA & GPT-4.1 & 60\% & 42\% & 26\% & \textbf{43\%} \\
DSPy & 32h & ReAct & w/ GEPA & GPT-4.1 & 30\% & 18\% & 4\% & 17\% \\
\hline
Koog & --- & SOTA & Hand-written agent & GPT-4.1 & 72\% & 56\% & 32\% & \textbf{53\%} \\
\hline
\end{tabular}
\caption{Performance of different agent optimization methods on Rocq proof generation, grouped by proof length (number of tactics). \emph{RB} denotes \textit{ReasoningBank}.}
\label{tab:agent-optimization-results}
\end{table*}

We study automatic optimization methods for agentic proof generation in Rocq and address two research questions. 
(i) Which automatic AI agent optimization methods lead to measurable performance gains over a fixed baseline agent in a production-oriented Rocq theorem proving setting? 
(ii) How does an automatically optimized proof agent compare to a state-of-the-art, human-engineered proof-agent pipeline in a Rocq setting?

We use the IMM theorem proving benchmark introduced in prior work on proof agents and reuse the dataset split released by CoqPilot~\citep{coqpilot}. The dataset contains 300 theorems from the IMM project~\citep{podkopaev2019bridging}. Theorems are split by difficulty using the length (in tactics) of the human-written reference proof as a proxy: shorter proofs tend to correspond to easier theorems, while longer proofs indicate harder instances. For evaluation, we select 50 theorems from each difficulty group. We allocate a separate set of 60 theorems for training the optimizers. Due to differences in experimental setup, the exact training subset used in the DSPy experiments slightly differs from the one used in the Koog experiments (the DSPy training split is mildly skewed toward easier theorems). This does not affect comparisons within each framework, where the training set is fixed, but may introduce minor variance when comparing results across frameworks. All experiments use GPT-4.1 due to its large context window and reliable tool use. 

For all experiments, we use the simplified RocqStar agent described in \Cref{sec:preliminaries} as the baseline architecture. We evaluate this agent in two implementations. In \emph{DSPy}, we consider a \emph{ReAct} strategy, which produces intermediate reasoning before making tool calls and serves as the basis for applying DSPy’s built-in optimization methods. In \emph{Koog}, we evaluate both a \emph{Simple} variant, which directly predicts tool calls without generating explicit reasoning traces, and a \emph{ReAct} variant, allowing us to assess whether explicit reasoning is necessary for effective optimization in this setting.

We train the optimization methods described in \Cref{sec:preliminaries} and evaluate them on the test set with a fixed resource budget of at most 20 tool calls. We report \emph{success rate} (percentage of theorems solved) for each difficulty group (A -- C) and overall (\textbf{Total}), as shown in \Cref{tab:agent-optimization-results}. We also report the \emph{optimization time}, i.e., the wall-clock time spent in the optimizer's training phase. 

In \Cref{tab:agent-optimization-results}, \emph{RB} denotes \emph{ReasoningBank}. Unless stated otherwise, ReasoningBank retrieves the top 5 memory items at test time based on encoder similarity and injects them into the agent context. The \emph{RB (no retrieval)} disables similarity-based retrieval and instead includes the entire memory bank in the context. For \emph{BootstrapFewShot}, we use at most four demonstrations due to context length constraints. We note that absolute performance differences between BootstrapFewShot in Koog and DSPy should be interpreted with caution, as they reflect minor framework-specific differences, including additional implicit prompting and tool-handling logic. \emph{GEPA} is run in its \emph{heavy} optimization mode, which enables full evolutionary search with reflective feedback, while \emph{MIPROv2} is run in the \emph{light} mode due to the substantially higher computational cost of the heavy configuration.%
\footnote{MIPROv2 with heavy mode is left for future experiments.}
For \emph{ADAS}, we perform ten optimization runs and report the best-performing agent obtained.

The experimental results reveal several patterns. Few-shot demonstrations lead to reliable performance improvements in all evaluated settings, making them the most robust optimization method in our study. When applied to agents that explicitly produce intermediate reasoning traces, as in the ReAct variants, few-shot demonstrations yield particularly strong gains. At the same time, few-shot demonstrations must be used with care, as longer traces can quickly exhaust the available context budget and substantially increase token usage. For context-centric methods, ReasoningBank benefits from retrieving a small number of relevant memory items rather than injecting the entire memory bank into the prompt, in contrast to ACE, which injects up to 200 memories into the context. This observation aligns with prior findings~\citep{Huang2025BreakingFC, xu2024rereadingimprovesreasoninglarge} that uncurated or weakly relevant context can degrade model performance by introducing noise. At the same time, effective use of ReasoningBank requires careful consideration of agent inputs: similarity-based retrieval assumes that the agent input is semantically meaningful for encoding, whereas in our setting the raw input initially consisted only of a theorem identifier and file path, with the actual proof state resolved later by the environment. As a result, contextual memory methods may require additional engineering to be applicable in theorem-proving pipelines.

Among prompt-centric optimizers, we find that more sophisticated methods do not consistently outperform simpler baselines in this setting. MIPROv2 is potentially well-suited for multi-component pipelines, where joint optimization of instructions and demonstrations across several agent nodes can be beneficial, but yields limited gains for the single-prompt agents studied here. SIMBA and MIPROv2 achieve improvements comparable to BootstrapFewShot in DSPy, suggesting that much of the observed performance gain stems from the inclusion of successful few-shot demonstrations rather than from deeper prompt evolution. Finally, ADAS, which attempts to optimize the agent’s control flow by generating executable agent code, performs inconsistently: despite multiple optimization iterations, we do not observe monotonic improvement, and the best-performing agent emerges early in the process. Moreover, the resulting agents exhibit strong bias toward the training data and rely on brittle, hard-coded decision logic, limiting their generality. 

Overall, none of the evaluated optimizers matches the performance of the hand-engineered state-of-the-art agent, indicating that fully automated optimization of agent architectures for formal verification remains an open challenge. Nevertheless, BootstrapFewShot emerges as a simple yet surprisingly strong optimization method, achieving consistent improvements with minimal engineering effort.

\section{Conclusion}
In this work, we evaluated a range of automatic agent optimization methods on a Rocq theorem proving task, using a simplified proof agent as the evaluation setting. Our results show that while several optimizers can yield measurable improvements over a fixed baseline, simple few-shot bootstrapping is the most consistently effective, especially for ReAct-style agents, improving overall success rates from 20\% to 33\% in Koog and from 19\% to 40\% in DSPy; however, none of the studied methods reaches the performance of a carefully engineered, state-of-the-art proof agent. Context-centric approaches such as ReasoningBank can be effective when contextual information is selectively retrieved, but sometimes require slight pipeline adaptations. More complex optimizers, including topology-level approaches such as ADAS, did not demonstrate consistent gains in this domain. Overall, our findings suggest that current automatic agent optimization techniques can partially reduce manual tuning effort but are not yet sufficient to fully automate the design of high-performing agentic systems for formal verification.

\bibliography{rocqsmith}

@inproceedings{lean,
  title={The Lean theorem prover (system description)},
  author={De Moura, Leonardo and Kong, Soonho and Avigad, Jeremy and Van Doorn, Floris and von Raumer, Jakob},
  booktitle={Automated Deduction-CADE-25: 25th International Conference on Automated Deduction, Berlin, Germany, August 1-7, 2015, Proceedings 25},
  pages={378--388},
  year={2015},
  organization={Springer},
  doi={https://doi.org/10.1007/978-3-319-21401-6_26}
}

@book{coq,
  title={Interactive theorem proving and program development: Coq’Art: the calculus of inductive constructions},
  author={Bertot, Yves and Cast{\'e}ran, Pierre},
  year={2013},
  publisher={Springer Science \& Business Media},
  doi={10.1007/978-3-662-07964-5}
}

@book{isabelle,
  title={Isabelle/HOL: a proof assistant for higher-order logic},
  author={Nipkow, Tobias and Wenzel, Markus and Paulson, Lawrence C},
  year={2002},
  publisher={Springer},
  doi={https://doi.org/10.1007/3-540-45949-9_5}
}

@inproceedings{compcert,
  title={CompCert-a formally verified optimizing compiler},
  author={Leroy, Xavier and Blazy, Sandrine and K{\"a}stner, Daniel and Schommer, Bernhard and Pister, Markus and Ferdinand, Christian},
  booktitle={ERTS 2016: Embedded Real Time Software and Systems, 8th European Congress},
  year={2016}
}

@InProceedings{yang2011finding,
  title={Finding and understanding bugs in C compilers},
  author={Yang, Xuejun and Chen, Yang and Eide, Eric and Regehr, John},
  booktitle={Proceedings of the 32nd ACM SIGPLAN conference on Programming language design and implementation},
  pages={283--294},
  year={2011},
  doi={https://doi.org/10.1145/1993498.1993532}
}

@inproceedings{coqpilot,
author = {Kozyrev, Andrei and Solovev, Gleb and Khramov, Nikita and Podkopaev, Anton},
title = {CoqPilot, a plugin for LLM-based generation of proofs},
year = {2024},
isbn = {9798400712487},
publisher = {Association for Computing Machinery},
address = {New York, NY, USA},
url = {https://doi.org/10.1145/3691620.3695357},
doi = {10.1145/3691620.3695357},
booktitle = {Proceedings of the 39th IEEE/ACM International Conference on Automated Software Engineering},
pages = {2382–2385},
numpages = {4},
location = {Sacramento, CA, USA},
series = {ASE '24}
}

@article{Huang2025BreakingFC,
  title={Breaking Focus: Contextual Distraction Curse in Large Language Models},
  author={Yue Huang and Yanbo Wang and Zixiang Xu and Chujie Gao and Siyuan Wu and Jiayi Ye and Xiuying Chen and Pin-Yu Chen and Xiangliang Zhang},
  journal={ArXiv},
  year={2025},
  volume={abs/2502.01609},
  url={https://api.semanticscholar.org/CorpusID:276107466}
}

@misc{xu2024rereadingimprovesreasoninglarge,
      title={Re-Reading Improves Reasoning in Large Language Models}, 
      author={Xiaohan Xu and Chongyang Tao and Tao Shen and Can Xu and Hongbo Xu and Guodong Long and Jian-guang Lou and Shuai Ma},
      year={2024},
      eprint={2309.06275},
      archivePrefix={arXiv},
      primaryClass={cs.CL},
      url={https://arxiv.org/abs/2309.06275}, 
}

@article{yang2023leandojo,
  title={Leandojo: Theorem proving with retrieval-augmented language models},
  author={Yang, Kaiyu and Swope, Aidan and Gu, Alex and Chalamala, Rahul and Song, Peiyang and Yu, Shixing and Godil, Saad and Prenger, Ryan J and Anandkumar, Animashree},
  journal={Advances in Neural Information Processing Systems},
  volume={36},
  pages={21573--21612},
  year={2023}
}

@article{agda,
title = {Programming language foundations in Agda},
journal = {Science of Computer Programming},
volume = {194},
pages = {102440},
year = {2020},
issn = {0167-6423},
doi = {https://doi.org/10.1016/j.scico.2020.102440},
url = {https://www.sciencedirect.com/science/article/pii/S0167642320300502},
author = {Wen Kokke and Jeremy G. Siek and Philip Wadler},
}

@article{podkopaev2019bridging,
  title={Bridging the gap between programming languages and hardware weak memory models},
  author={Podkopaev, Anton and Lahav, Ori and Vafeiadis, Viktor},
  journal={Proceedings of the ACM on Programming Languages},
  volume={3},
  number={POPL},
  pages={1--31},
  year={2019},
  publisher={ACM New York, NY, USA}
}

@misc{kozyrev2025rocqstar,
  title={RocqStar: Leveraging Similarity-driven Retrieval and Agentic Systems for Rocq generation}, 
  author={Andrei Kozyrev and Nikita Khramov and Gleb Solovev and Anton Podkopaev},
  year={2025},
  eprint={2505.22846},
  archivePrefix={arXiv},
  primaryClass={cs.LG},
  url={https://arxiv.org/abs/2505.22846}, 
}

@inproceedings{nlir,
  title = {NLIR: Natural Language Intermediate Representation for Mechanized Theorem Proving},
  author = {Teodorescu, Laetitia and Baudart, Guillaume and Gallego Arias, Emilio Jes{\'u}s and Lelarge, Marc},
  url = {https://hal.science/hal-04886208},
  booktitle = {{MathAI@NeuRIPS 2024 - 4th Workshop on Mathematical Reasoning and AI}},
  address = {Vancouver, Canada},
  year = {2024},
  month = Dec,
  hal_id = {hal-04886208},
  hal_version = {v1},
}

@misc{ace,
  title={Agentic Context Engineering: Evolving Contexts for Self-Improving Language Models}, 
  author={Qizheng Zhang and Changran Hu and Shubhangi Upasani and Boyuan Ma and Fenglu Hong and Vamsidhar Kamanuru and Jay Rainton and Chen Wu and Mengmeng Ji and Hanchen Li and Urmish Thakker and James Zou and Kunle Olukotun},
  year={2025},
  eprint={2510.04618},
  archivePrefix={arXiv},
  primaryClass={cs.LG},
  url={https://arxiv.org/abs/2510.04618}, 
}

@misc{reasoningbank,
  title={ReasoningBank: Scaling Agent Self-Evolving with Reasoning Memory}, 
  author={Siru Ouyang and Jun Yan and I-Hung Hsu and Yanfei Chen and Ke Jiang and Zifeng Wang and Rujun Han and Long T. Le and Samira Daruki and Xiangru Tang and Vishy Tirumalashetty and George Lee and Mahsan Rofouei and Hangfei Lin and Jiawei Han and Chen-Yu Lee and Tomas Pfister},
  year={2025},
  eprint={2509.25140},
  archivePrefix={arXiv},
  primaryClass={cs.AI},
  url={https://arxiv.org/abs/2509.25140}, 
}

@misc{mipro,
  title={Optimizing Instructions and Demonstrations for Multi-Stage Language Model Programs}, 
  author={Krista Opsahl-Ong and Michael J Ryan and Josh Purtell and David Broman and Christopher Potts and Matei Zaharia and Omar Khattab},
  year={2024},
  eprint={2406.11695},
  archivePrefix={arXiv},
  primaryClass={cs.CL},
  url={https://arxiv.org/abs/2406.11695}, 
}

@misc{adas,
      title={Automated Design of Agentic Systems}, 
      author={Shengran Hu and Cong Lu and Jeff Clune},
      year={2025},
      eprint={2408.08435},
      archivePrefix={arXiv},
      primaryClass={cs.AI},
      url={https://arxiv.org/abs/2408.08435}, 
}

@misc{react,
      title={ReAct: Synergizing Reasoning and Acting in Language Models}, 
      author={Shunyu Yao and Jeffrey Zhao and Dian Yu and Nan Du and Izhak Shafran and Karthik Narasimhan and Yuan Cao},
      year={2023},
      eprint={2210.03629},
      archivePrefix={arXiv},
      primaryClass={cs.CL},
      url={https://arxiv.org/abs/2210.03629}, 
}

@misc{gepa,
  title={GEPA: Reflective Prompt Evolution Can Outperform Reinforcement Learning}, 
  author={Lakshya A Agrawal and Shangyin Tan and Dilara Soylu and Noah Ziems and Rishi Khare and Krista Opsahl-Ong and Arnav Singhvi and Herumb Shandilya and Michael J Ryan and Meng Jiang and Christopher Potts and Koushik Sen and Alexandros G. Dimakis and Ion Stoica and Dan Klein and Matei Zaharia and Omar Khattab},
  year={2025},
  eprint={2507.19457},
  archivePrefix={arXiv},
  primaryClass={cs.CL},
  url={https://arxiv.org/abs/2507.19457}, 
}

@article{fu2025weaknessesofcoqpilot,
    author = {Fu, Yujia and Liang, Peng and Tahir, Amjed and Li, Zengyang and Shahin, Mojtaba and Yu, Jiaxin and Chen, Jinfu},
    title = {Security Weaknesses of Copilot-Generated Code in GitHub Projects: An Empirical Study},
    year = {2025},
    issue_date = {November 2025},
    publisher = {Association for Computing Machinery},
    address = {New York, NY, USA},
    volume = {34},
    number = {8},
    issn = {1049-331X},
    url = {https://doi.org/10.1145/3716848},
    doi = {10.1145/3716848},
    journal = {ACM Trans. Softw. Eng. Methodol.},
    month = oct,
    articleno = {218},
    numpages = {34}
}

@misc{dspy,
  title={DSPy: Compiling Declarative Language Model Calls into Self-Improving Pipelines}, 
  author={Omar Khattab and Arnav Singhvi and Paridhi Maheshwari and Zhiyuan Zhang and Keshav Santhanam and Sri Vardhamanan and Saiful Haq and Ashutosh Sharma and Thomas T. Joshi and Hanna Moazam and Heather Miller and Matei Zaharia and Christopher Potts},
  year={2023},
  eprint={2310.03714},
  archivePrefix={arXiv},
  primaryClass={cs.CL},
  url={https://arxiv.org/abs/2310.03714}, 
}

@misc{bayesianoptimization,
  title={Practical Bayesian Optimization of Machine Learning Algorithms}, 
  author={Jasper Snoek and Hugo Larochelle and Ryan P. Adams},
  year={2012},
  eprint={1206.2944},
  archivePrefix={arXiv},
  primaryClass={stat.ML},
  url={https://arxiv.org/abs/1206.2944}, 
}

@misc{xu2026agentic,
      title={Agentic Proof Automation: A Case Study}, 
      author={Yichen Xu and Martin Odersky},
      year={2026},
      eprint={2601.03768},
      archivePrefix={arXiv},
      primaryClass={cs.PL},
      url={https://arxiv.org/abs/2601.03768}, 
}
\bibliographystyle{iclr2026_conference}

\end{document}